\documentclass[10pt,conference]{IEEEtran} 
\usepackage[backend=bibtex]{biblatex} %Imports biblatex package
\usepackage{multirow}
\usepackage[tableposition=top]{caption}
\usepackage{xcolor}
\usepackage[many]{tcolorbox}
\usepackage{amsmath}
\usepackage{amssymb, subcaption, caption}
\usepackage{graphicx}
\usepackage{hyperref}
\renewcommand{\footnotesize}{\scriptsize} 
\bibliography{reference.bib} 

\begin{document}
\title{Robustness Evaluation in Hand Pose Estimation Models using Metamorphic Testing}
\makeatletter
\newcommand{\linebreakand}{%
  \end{@IEEEauthorhalign}
  \hfill\mbox{}\par
  \mbox{}\hfill\begin{@IEEEauthorhalign}
}

\author{
\IEEEauthorblockN{Muxin Pu\IEEEauthorrefmark{1}, Chun Yong Chong\IEEEauthorrefmark{1}, Mei Kuan Lim\IEEEauthorrefmark{1}}
\IEEEauthorblockA{\IEEEauthorrefmark{1}School of Information Technology,\\ 
Monash University Malaysia\\
\{muxin.pu, chong.chunyong, lim.meikuan\}@monash.edu}
}

% \author{\IEEEauthorblockN{Muxin Pu}
% \IEEEauthorblockA{\textit{School of Information Technology} \\
% \textit{Monash University Malaysia}\\
% Malaysia \\
% muxin.pu@monash.edu}
% \and
% \IEEEauthorblockN{Chun Yong Chong}
% \IEEEauthorblockA{\textit{School of Information Technology} \\
% \textit{Monash University Malaysia}\\
% Malaysia \\
% chong.chunyong@monash.edu}
% \and
% \IEEEauthorblockN{Mei Kuan Lim}
% \IEEEauthorblockA{\textit{School of Information Technology} \\
% \textit{Monash University Malaysia}\\
% Malaysia \\
% lim.meikuan@monash.edu}
% }

\maketitle
\begin{abstract}
Hand pose estimation (HPE) is a task that predicts and describes the hand poses from images or video frames. When HPE models estimate hand poses captured in a laboratory or under controlled environments, they normally deliver good performance. However, the real-world environment is complex, and various uncertainties may happen, which could degrade the performance of HPE models. For example, the hands could be occluded, the visibility of hands could be reduced by imperfect exposure rate, and the contour of hands prone to be blurred during fast hand movements. In this work, we adopt metamorphic testing to evaluate the robustness of HPE models and provide suggestions on the choice of HPE models for
different applications. The robustness evaluation was conducted on four state-of-the-art models, namely MediaPipe hands, OpenPose, BodyHands, and NSRM hand. We found that on average more than 80\% of the hands could not be identified by BodyHands, and at least 50\% of hands could not be identified by MediaPipe hands when diagonal motion blur is introduced, while an average of more than 50\% of strongly underexposed hands could not be correctly estimated by NSRM hand. Similarly, applying occlusions on only four hand joints will also largely degrade the performance of these models. The experimental results show that occlusions, illumination variations, and motion blur are the main obstacles to the performance of existing HPE models. These findings may pave the way for researchers to improve the performance and robustness of hand pose estimation models and their applications. 
\end{abstract}


\IEEEpeerreviewmaketitle

\section{Introduction}
Hand pose estimation (HPE) has drawn increasing attention from the computer vision research communities due to its promising application in various domains such as sign language recognition \cite{john2022hand}, gesture recognition \cite{zhang2020mediapipe}, virtual, or augmented reality \cite{4480766}. In general, HPE consists of two components, namely hand segmentation and hand landmark localisation \cite{grzejszczak2016hand}. The former involves segmenting hands from an image or video frame, while the latter maps image features to hand pose parameters. HPE is essentially a technique of capturing a set of coordinates for each joint, known as \textbf{key points} \cite{chen2020nonparametric, simon2017hand}, that can describe a hand pose. To facilitate intuitive and natural interaction, various interaction styles tend to incorporate both modalities and various learning-based approaches have been utilised in the literature which have shown excellent results \cite{chen2020survey, erol2007vision}. 

Regardless of the fast development, it is undeniable that learning-based models frequently fail in unintuitive ways.  
One of the greatest challenges for learning-based models is domain adaptation \cite{zhou_2022}, which requires HPE models to withstand a wide variety of operating conditions and handle unfamiliar datasets. % Domain adaptation, the ability of HPE models to withstand a wide variety of operating conditions and handle unseen datasets, is one of the greatest challenges \cite{zhou_2022}.%
Unlike face and body, large datasets of annotated key points for hands are rare \cite{chen2020nonparametric}. The quality of HPE models is reliant on the availability of training data; such data are typically obtained in controlled environments with little variation \cite{FreiHand2019}. The ability of these models to handle unfamiliar domains may be insufficient due to domain distribution discrepancies \cite{zhou_2022}, meaning that most source and target data are gathered from the same probability distribution. This could restrict the actual application of HPE models or, to some degree, cause these models to perform well in lab settings but poorly in practice. 

To encourage the use of HPE models in practice, it is necessary to fulfil the wide settings of real-world scenarios \cite{chen2020nonparametric, myanganbayar2018partially, zhang2020mediapipe}. However, the existing literature has found that the performance and robustness of HPE models suffer from occlusion \cite{erol2007vision, khaleghi2022multi, zhao2020perceiving}, illumination variations \cite{khaleghi2022multi}, and motion blur \cite{8998145}, which are common in practice. %The likelihood a key point on the hand is self-occlusion by the hand or occlusion by objects is considerably high.% 
Key points of hands are prone to be occluded or self-occluded due to articulation, viewpoint, or grasped objects during gesturing or object interactions. Besides, the changes in visibility by illumination variations are also unavoidable. Motion blur caused by fast hand movement occurs commonly as well. The aforementioned settings are the main hurdles commonly faced by HPE models. Unfortunately, their impact on HPE models have not been well investigated in the existing literature, and there are limited qualitative measurements made specifically towards them. 

% To encourage the use of HPE models in practice, it is necessary to fulfil the wide set of characteristics of real-world scenarios \cite{zhang2020mediapipe, chen2020nonparametric, myanganbayar2018partially}, such as varying illumination, occlusion level, and motion blur. The aforementioned characteristics are the main hurdles commonly faced by HPE models but unfortunately have not been well investigated in the existing literature. 

One of the main reasons why these operating constraints/settings (illumination, occlusion, etc.) are not explored thoroughly in the extant studies is due to the lack of a test oracle. Existing HPE-related open-source datasets \cite{simon2017hand, FreiHand2019} usually consist of hand images or videos under pristine conditions with no artefacts, illumination variations, or motion blur, making it difficult for researchers to evaluate the robustness of their proposed model in a less than ideal operating environment.

On the other hand, testing HPE models is challenging, since learning-based models are built on enormous input spaces with probabilistic results from largely non-transparency components rather than explicitly defined inputs and logical flows based on explicit programming statements.

Metamorphic testing (MT) has been widely used to test learning-based models in the test oracle problem due to its simplicity of concept and effectiveness in fault detection \cite{chen2018metamorphic, lim2022metamorphic, pu2022fairness, 9036058}. MT verifies and validates the robustness of models against metamorphic relations (MRs), which are essential properties of the target algorithms or models in relation to various inputs and their expected outputs \cite{chen2018metamorphic}. The first step of MT implementation involves generating or obtaining source inputs as source test cases. MRs are then used to derive follow-up test cases from the source input. Unlike the traditional way of evaluating each test case separately, MT verifies the correctness of the source and follow-up test cases by comparing their output with the corresponding MR \cite{chen2018metamorphic}.

Motivated by the demand for robust HPE models that can be widely used in real applications, we use MT to evaluate the robustness of HPE models. The specific goals of this study are as follows:
\begin{itemize}
    \item To propose metamorphic relations for evaluating the robustness of HPE models in uncontrolled settings that mimic the real world deployment environment. %with variations in occlusion level, illumination, and motion blur. 
    \item To provide a thorough analysis of the effect of these settings on the performance of state-of-the-art HPE models.
    \item To provide suggestions on the choice of HPE models for different applications based on the strength and vulnerability of each HPE model. 
\end{itemize}

\section{Related Work}
\textbf{Hand pose estimation: }Vision-based hand pose estimation (HPE) has been well studied in the literature for years. Numerous techniques have been proposed, and there appears to be no dominant method in this field. Every year, the performance gains are shown by both the traditional optimisation approach based on the proposed algorithm and a data-driven approach, such as deep learning. The work by Tompson et al. \cite{tompson2014real} proposed the feedback loop architecture of a convolutional neural network, which shows an outstanding result with excellent efficiency. The proposed approach is a data-driven approach with the goal of iteratively correcting errors. In the work by Simon et al. \cite{simon2017hand}, they present an approach called multiview bootstrapping, where a multi-camera system is used to train fine-grained detectors. In addition, MediaPipe hands \cite{zhang2020mediapipe} utilizes a machine-learning pipeline consisting of a palm detector and a hand landmark model that works together. MediaPipe hands model achieves an outstanding result in real-time on-device HPE. Their experiments have shown that MediaPipe hands is robust even to occluded hands. 

%for key points that are prone to occlusion. Moreover, Chen et al. \cite{chen2020nonparametric} introduced a novel cascade structure regularization methodology. The qualitative results presented in \cite{chen2020nonparametric} demonstrate that their model effectively improves the consistency of the structure to estimate the pose of the hand, especially when there is severe occlusion. 

\textbf{Robustness evaluation: }There are relatively smaller numbers of research that focus on evaluating pose estimation models, compared to image classification and object detection. Dollar et al. \cite{dollar2010cascaded} discovered that, unlike human annotators, algorithms have a bimodal distribution of normalised distances between a part detection and the ground truth, indicating multiple error modes. In \cite{ruggero2017benchmarking}, the authors introduce a novel method to analyse errors in multi-instance pose estimation algorithms and a principled benchmark for comparing different algorithms. In their work, three types of errors are defined and  characterised: localisation, scoring, and background. The authors investigate how instance attributes affect these errors and algorithm performance. However, to the best of our knowledge, existing works have not been evaluated thoroughly on scenarios that could potentially affect the HPE models' performance. There are various scenarios where different degrees of occlusion, varying illumination, and motion blur might be involved, which are very likely to occur in real applications. Thus, it is essential to understand further how these models behave under such scenarios to be used in practice.  

\textbf{Metamorphic testing: }Metamorphic testing (MT) has been used in various application domains to detect faults, and it has also been integrated with other software analysis and testing technologies to extend its applicability to systems with or without a test oracle \cite{park2021robustness}. For example, MT managed to identify new faults \cite{Rao2013, xie2013866} in 3 out of 7 programs in the Siemens suite \cite{Hutchins1994}, a software for drivers, although this application had been studied repetitively in major software testing research projects for two decades. Furthermore, Le et al. \cite{le2014compiler} use a simple MR to identify more than 100 flaws in two popular C compilers (GCC and LLVM). In addition to its application in software testing, MT has been broadly considered a mainstream and promising approach for handling oracle problems in the wider context of software engineering. It has also been used as a validation and quality assessment approach to detect real-world flaws in many popular learning-based systems \cite{chen2018metamorphic}. 

\section{Motivation}
According to the current research on HPE, most HPE models are trained and tested in ideal environments, which may not reflect the wide range of real operations. However, we generally expect a robust model that performs equally well in both controlled and uncontrolled operating environments. Therefore, it is necessary to expose HPE models to different kinds of noises that could mimic real operating environments in order to evaluate the robustness of HPE models when confronted with such problems. Given the effectiveness of MT in both test case generation and fault detection, our work primarily focuses on utilizing MT as a tool to produce test cases and evaluate the performance and robustness of HPE models under uncontrolled settings that simulate real-world operating environments.

\section{Methods}

We start our evaluation with source test cases that contain images directly sampled from the original datasets \cite{simon2017hand, FreiHand2019} without any transformation, where two types of source test cases are involved for hands with and without objects. To have a clear observation on the influence of individual variation (MRs), we only involve both of the source test cases in the first place to obtain a reliable baseline and verify each MR by applying the transformation only to images containing hands without objects. Evaluation metrics (refer to Section \ref{Experiments}) are calculated for each test result to assess the robustness of selected HPE models. These metrics are then used to determine if MRs are satisfied or violated. 

% \subsection{Metamorphic relations}
As a primary consideration, we transform the source test case (or original images) and construct the corresponding follow-up test cases consistent with the requirements to evaluate the robustness of HPE models, including occlusion, exposure, and motion blur, which are common in the actual operating environment. Considering the influence of uncontrolled settings, we propose the following MRs \footnote{ \url{https://github.com/mpuu00001/Robustness-Evaluation-in-Hand-Pose-Estimation}\newline We provide a summary of all MRs and the derived TCs on our GitHub page.}: 
\begin{itemize}
    \item \textbf{MR{$_1$}} is based on the expectation to observe a strong correlation between the performance of HPE models and the number of occluded key points of the target hands.
    \item \textbf{MR{$_2$}} assumes that the performance of HPE models should not be degraded significantly by the introduction of slight occlusion on hand regions, that are prone to be occluded, such as a particular finger.
    % Should not degrade its performance when common key points such as fingers or palms are occluded
    % When common key points are occluded.
    % In a natural setting, common key points can include a particular finger or palms, as depicted in Figure xxx
    \item \textbf{MR{$_3$}} hypothesize that if we change the exposure rate of the original image, which mimic the illumination of different operating environments, the performance of HPE models should not be degraded. 
    \item \textbf{MR{$_4$}} expects that the performance of HPE models should not be degraded by the introduction of motion blur, which simulate real-world fast hand movement.
\end{itemize} 

The strong correlation mentioned in MR{$_1$} means we expect a high degree of consistency between the performance of HPE models and the number of occluded key points. 

\textbf{MR{$_1$}} $\&$ \textbf{MR{$_2$}}: We built these two MRs based on the observation that occlusion is unavoidable in practice. Thus, reliable HPE models should show robust performance when occlusion is introduced. In the MS-COCO 2017 validation dataset \cite{ruggero2017benchmarking}, a total of 17 key points are used to annotate a body pose, and the authors classify body poses with 1 to 6 occluded key points as slightly occluded. Based on this, the introduction of 4 occluded key points is referred to the introduction of slight occlusion in our study, where a hand landmark of 21 key points is used to annotate a hand pose (see Figure \ref{fig:key_points}). Moving on, the test image $i'_1$ for MR{$_1$} and MR{$_2$} are obtained by adding noise to occlude the target hand in its original image $i$. Then, given the occlusion artifact $o$, the coordinates of hand landmarks $H$, and the corresponding indices $X$ that match the occlusion artifact with its targeted hand landmark, the test image $i'_1$ for MR{$_1$} and MR{$_2$} can be expressed as: 
\begin{equation} \label{eq:i'1}
    i'_1 = {h^x(i, o), h \in H \ \mathrm{and}\ x \in X}
\end{equation}
The test image $i'_1$ is then generated by iteratively applying the occlusion artifact $o$ on its original image $i$ at each key point identified by $h^x$. 

\begin{figure}[h!]
    \centering
    \includegraphics[width=0.25\textwidth]{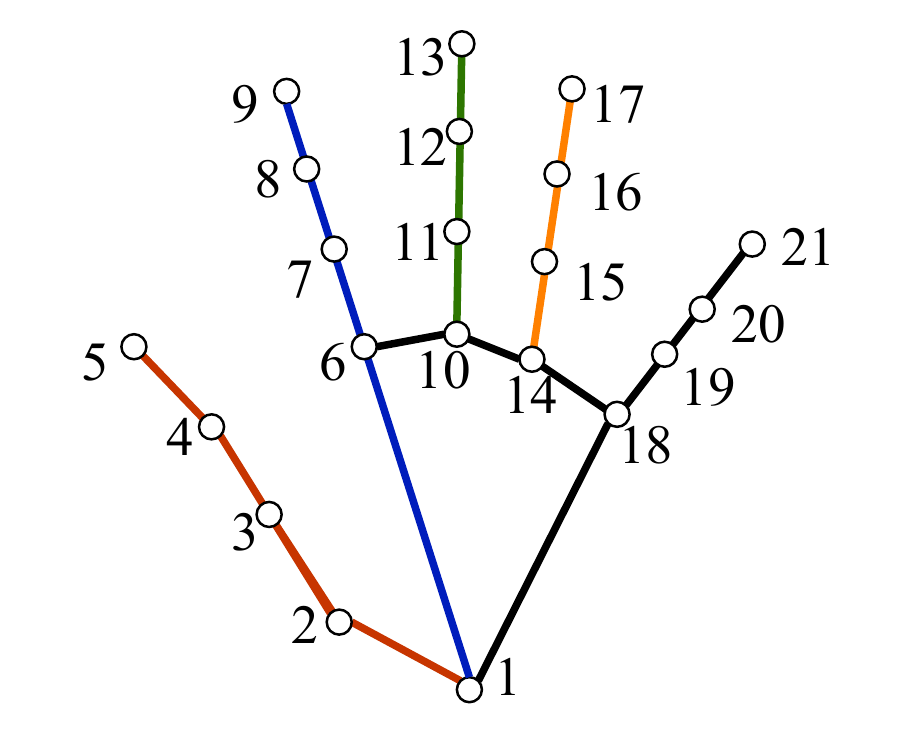}
    \caption{The hand landmark of 21 key points.}
    \label{fig:key_points}
\end{figure}

\textbf{MR{$_3$}}: The applicability of HPE models will be limited if they only work under controlled lighting conditions. Hence, MR{$_3$} is built to gauge the robustness of the models under such a variation. The test image $i'_2$ for MR{$_3$} is obtained by changing the gamma value $\gamma$ of its original image $i$. The pixel intensities of $i$ are first scaled from the range [0, 255] to [0, 1]. From there, we generate the test image $i'_2$ using the following function: 
\begin{equation} \label{eq:i'2}
    i'_2 = {i^{\gamma}}
\end{equation}
The test image $i'_2$ is then obtained by scaling back to the range of [0, 255]. Moreover, $\gamma > 1$ will shift the original image towards the darker end of the spectrum while $\gamma < 1$  will make the image appear lighter, and $\gamma = 1$ will have no effect on the original image. In order to simulate different lighting conditions, we use four different gamma values to generate test cases including strong underexposure, underexposure, overexposure, and strong overexposure, where the gamma values are within a reasonable range between 0 to 5.5. 

\textbf{MR{$_4$}}: When we gesture, interact with objects, or manipulate devices with our hands, motion blur will occur if they are captured by cameras. This common phenomenon inspires us to derive MR{$_4$}. The test image $i'_3$ for MR{$_4$} is obtained by adding a correlation kernel or filter to the original image $i$. % The basic step of applying correlation filtering is as follows: 1) Put the kernel anchor on the top of a determined pixel, with the rest of the kernel overlaying the pixels of the original image. 2) Multiply the kernel coefficients by the matched pixels and then sum up the result. 3) Return the result and place it at the location of the anchor in the original image. 4) Scan the kernel over the image and repeat the procedure for all pixels. 
Then, given a correlation kernel $k$ of size $m \times m$, and kernel anchor $a$ that indicates the relative location of a filtered point within the kernel. The test image $i'_3$ can be expressed as: 
\begin{equation} \label{eq:i'3}
    i'_3(x', y') = \sum_{x=1} ^{m_x-1} \sum_{y=1}^{m_y-1} i(x+x'-a_x, y+y-a_y)k(x, y)
\end{equation}
% Where a kernel anchor $a = (-1, -1)$ is used to produce the test image $i'_3$, which means the anchor is at the kernel centre. 
Where a fixed filter point indicated by $a$ is used and located at the centre of the correlation kernel. The kernel $k$ contains only 1's and 0's. The direction of 1’s across the kernel grid is the direction of the desired motion. Upon this, we produce three types of motion blur, including horizontal, vertical, and diagonal motion blur using the same kernel size of 20 $\times$ 20. The size is of reasonable for testing images of size 244 $\times$ 244. 

\section{Experiments} \label{Experiments}
\subsection{Hand Pose Estimation Models}
%MediaPipe hands \cite{zhang2020mediapipe}, OpenPose \cite{simon2017hand}, BodyHands \cite{narasimhaswamy2022whose}, and NSRM hand \cite{chen2020nonparametric} are selected as the HPE models for our robustness evaluation. 
% In this study, we directly use the pre-trained models provided by the original authors on their official GitHub repositories. 
The four chosen state-of-the-art models provide different solutions to HPE problems. MediaPipe hands is a standalone HPE model that is capable of both hand segmentation and hand landmark localization. In contrast, OpenPose requires predefined hand boundary boxes and BodyHands is designed specifically for hand segmentation, and hence, we integrate them together to form a comprehensive HPE model. The hand area segmented by BodyHands will then be used by OpenPose for hand landmark localisation. NSRM hand, on the other hand, assumes that there is always a hand present and requires testing images aligning with the hands. As such, a prepossessing step is involved to meet the input requirement for NSRM hand model.  

\subsection{Datasets}
We have chosen two public hand pose datasets, namely FreiHand \cite{FreiHand2019} and CMU Panoptic Hand \cite{simon2017hand} (Panoptic), where both contain real-captured images with various hand poses associated with their key point coordinate annotations. Since we directly use the pre-trained models, a subset of data is selected and used only for testing purposes. Testing images are then classified into hands with and without objects to avoid unfair comparisons between the two datasets. As a result, we select 2500 sample images containing hands with objects and 3000 sample images containing hands without objects for each dataset used in this study. Additionally, we did not involve any resizing or cropping for the images from the FreiHand dataset, since all the images are occupied by one hand and are of a unified size of 244 $\times$ 244. As for the Panoptic dataset, images retrieved from the dataset were cropped by Chen et. al. \cite{chen2020nonparametric} based on a square patch, which is of size 2.2 times the largest dimension of the tightest bounding box enclosing all hand key points. We then resized these images to be the same size as those from the FreiHand dataset. 

\subsection{Evaluation Metrics}
% It is critical for us to determine individual instances of errors in the models. As such, 

Confusion matrices assist us in visualising the various sets of predictions and calculating the precision, recall, and F1-score for each test case. Hand segmentation boundaries outputted from the HPE models are measured against the ground truth boundaries obtained from the selected datasets using intersection over union (IoU). The IoU values are used to assess the overlapping between two bounding boxes with the same prediction label. In our case, the boundary box means the tightest box enclosing all hand key points and a threshold of IoU = 0.5 is commonly suggested \cite{solovyev2021weighted} to separate prediction results. For hand landmark localization, Euclidean distance (ED) is calculated between key points to measure how well the models describe a hand pose. According to FreiHand \cite{FreiHand2019}, they use two distances, 5 mm and 15 mm, as the threshold to evaluate the performance of the hand landmark localization. Therefore, it is reasonable to take the mean of these two distances and set 10 mm as our threshold. 

Next, we define the positive outcomes derived from the HPE models as shown in Table \ref{tab:positive_outcomes}. Positive outcomes indicate there are predictions provided by the models, while negative outcomes indicate the models did not provide any predictions. Each prediction is then assessed as true or false to tell whether it is a correct or incorrect prediction. Note that all the testing images are occupied by one hand, so there is no negative test case leading to a true negative outcome.

\begin{table}[h!]
    \centering
    \caption{The positive outcomes derived from the selected HPE models}
    \label{tab:positive_outcomes}   
    \resizebox{0.478\textwidth}{!}{%
    \begin{tabular}{lll}
    \hline
    Categories & True positive & False positive \\ \hline
    Hand segmentation & IoU \textgreater 0.5 & 0 \textgreater IoU \textless{}= 0.5 \\
    Hand landmark localisation & ED \textless 10 & ED \textgreater{}= 10 \\ \hline
    \end{tabular}%
    }
\end{table}

After the thresholds are determined and confusion matrices are constructed taking into account all prediction results, the precision, recall, and the F1-score can be further derived. Here, precision is chosen as a measure of quality, while recall is a measure of quantity. A high precision indicates that an algorithm returns more relevant results than irrelevant ones. In our case, it assesses how precise the identified hand poses are estimated. A high recall indicates that an algorithm is lenient in its criteria for identifying positive predictions. It measures how sensitive the model is to hands. F1-score is the harmonic mean of precision and recall, which accounts for both false positives and false negatives. These measures offer us a deeper insight into the estimation made by the selected HPE models and the transformation in the input-output pairs of the models. 

\subsection{Results and Discussion}
Prior to verify any metamorphic relations, the original datasets are classified into two categories for hands with and without any object (refer to Figure \ref{fig:source_test_cases}). %with the sake of alleviating unfair comparisons between the FreiHand and Panoptic datasets.%
We then start our experiments by testing the pre-trained models on the classified datasets to obtain a reliable baseline (see Table \ref{tab:results_hand_without_obj}). 

\begin{figure}[t]
    \centering
    \captionsetup{justification=centering}
    \begin{subfigure}[b]{0.14\textwidth}
        \caption*{Without any object}
        \includegraphics[width=\textwidth]{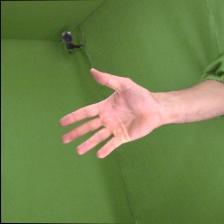}
    \end{subfigure}
    \hspace{3em}
    \begin{subfigure}[b]{0.14\textwidth}
        \caption*{With an object}
        \includegraphics[width=\textwidth]{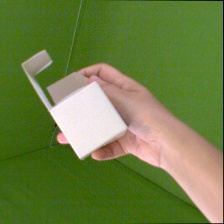}
    \end{subfigure}
    \caption{Samples of images in the classified original datasets}
    \label{fig:source_test_cases}
\end{figure}

\begin{table*}[t]
    \centering
    \caption{Results by original images of \textbf{hands without any objects} (\textbf{baseline})}
    \resizebox{0.85\textwidth}{!}{%
    \begin{tabular}{cccccccc}
    \hline
    \multirow{2}{*}{Model} & \multirow{2}{*}{Dataset} & \multicolumn{3}{c}{Hand segmentation result} & \multicolumn{3}{c}{Hand landmark localisation result} \\ \cline{3-8} 
     &  & Precision\% & Recall\% & F1-score\% & Precision\% & Recall\% & F1-score\% \\ \hline
    \multirow{2}{*}{MediaPipe hands} & FreiHand & 97.59\% & 95.33\% & 96.44\% & 98.69\% & 95.38\% & 97.00\% \\
     & Panoptic & 94.69\% & 73.59\% & 82.81\% & 97.59\% & 74.17\% & 84.28\% \\ \hline
    \multirow{2}{*}{BodyHands+Openpose} & FreiHand & 99.42\% & 80.27\% & 88.83\% & 98.59\% & 80.14\% & 88.41\% \\
     & Panoptic & 98.34\% & 81.92\% & 89.38\% & 97.92\% & 81.86\% & 89.17\% \\ \hline
    \multirow{2}{*}{NSRM hand} & FreiHand & 88.60\% & 100.00\% & 93.96\% & 84.39\% & 100.00\% & 91.54\% \\
     & Panoptic & 99.87\% & 100.00\% & 99.93\% & 99.78\% & 100.00\% & 99.89\% \\ \hline
    \end{tabular}%
        }
    \label{tab:results_hand_without_obj}
\end{table*}

\begin{table*}[t]
    \centering
    \caption{Results by original images of \textbf{hands with objects}}
    \resizebox{0.85\textwidth}{!}{%
    \begin{tabular}{cccccccc}
    \hline
    \multirow{2}{*}{Model} & \multirow{2}{*}{Dataset} & \multicolumn{3}{c}{Hand segmentation result} & \multicolumn{3}{c}{Hand landmark localisation result} \\ \cline{3-8} 
     &  & Precision\% & Recall\% & F1-score\% & Precision\% & Recall\% & F1-score\% \\ \hline
    \multirow{2}{*}{MediaPipe hands} & FreiHand & 95.68\% & 85.67\% & 90.40\% & 97.41\% & 85.88\% & 91.29\% \\
     & Panoptic & 88.42\% & 51.86\% & 65.37\% & 93.40\% & 53.23\% & 67.81\% \\ \hline
    \multirow{2}{*}{BodyHands+Openpose} & FreiHand & 97.40\% & 84.34\% & 90.40\% & 96.41\% & 84.20\% & 89.89\% \\
     & Panoptic & 43.98\% & 70.85\% & 54.27\% & 46.51\% & 71.99\% & 56.51\% \\ \hline
    \multirow{2}{*}{NSRM hand} & FreiHand & 83.96\% & 100.00\% & 91.28\% & 78.59\% & 100.00\% & 88.01\% \\
     & Panoptic & 41.12\% & 100.00\% & 58.28\% & 41.30\% & 100.00\% & 58.45\% \\ \hline
    \end{tabular}%
        }
    \label{tab:results_hand_with_obj}
\end{table*}

By comparing Table \ref{tab:results_hand_without_obj} and Table \ref{tab:results_hand_with_obj}, we can observe that the results obtained from the Panoptic dataset generally decrease more drastically than those obtained from the FreiHand dataset.  
%Based on the results, we compute the mean differences between them in Table \ref{tab:mean_difference}, where the results obtained from the Panoptic dataset show a drastic performance degradation. 
This degradation could be caused by multiple reasons, such as fast hand movements, illumination variations, occlusions, and information loss during image resizing or cropping. To understand the cause of the poor performance, 557 testing images are filtered out from the Panoptic dataset and a new round of testing is conducted based on it, which boosts their performance and gives us ideas on the robustness evaluation with respect to the aforementioned settings. The detailed results can be found in our GitHub page.

% The selected models achieve relatively good performance on the sub-sample datasets that consist of images of hands without any object. They have some degree of performance degradation when tested 
On both datasets, all of these models exhibit a similar pattern, which is a decline in their performance when tested on unfamiliar domains (testing on images that contain severely occluded hands with the introduction of foreign objects on the focal hand). 
% Although Openpose and NSRM hand are trained on the Panoptic dataset, they still fail to adequately gain useful information to handle occlusion problems. 
Consequently, the multiple modes presented by the occluded joints still remain as an unfamiliar domain for them. From these results, we can also deduce that occlusions can be an obstacle to the performance of HPE models and none of these models is robust enough to deal with severely occluded hands. 

% We are then interested to evaluate the relationship between occlusions and the performance of HPE models by MR{$_1$} and the corresponding test case TC{$_1$}. 
We then proceed with evaluating the relationship between occlusions and the performance of HPE models by MR{$_1$} via test case TC{$_1$}, which consists of 21 sub-groups and the number of occluded key points increases with the level of occlusion. Figure \ref{fig:MR1_samples} shows samples of testing images in TC{$_1$}. For level $n$ occlusion test cases, black circles 10 pixels in radius are introduced at key points indexed by the first $n$ joints of the hand landmarks provided by the original database. As stated in MR{$_1$}, we expect to observe a strong linear correlation between the performance of HPE models and the number of occluded key points, and thus, each key point is equally important for hand pose estimation. 

\begin{figure}[b]
    \centering
    \begin{subfigure}[b]{0.14\textwidth}
        \caption*{Level 1 occlusion}
        \includegraphics[width=\textwidth]{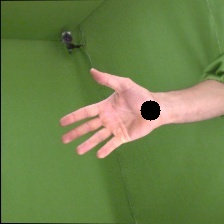}
    \end{subfigure}
    ~
    \begin{subfigure}[b]{0.14\textwidth}
        \caption*{Level 3 occlusion}
        \includegraphics[width=\textwidth]{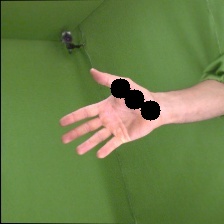}
    \end{subfigure}
    ~
    \begin{subfigure}[b]{0.14\textwidth}
        \caption*{Level 6 occlusion}
        \includegraphics[width=\textwidth]{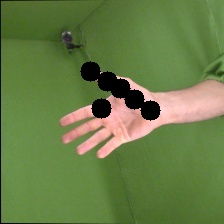}
    \end{subfigure}
    \caption{Samples of images in TC{$_1$}}
    \label{fig:MR1_samples}
\end{figure}

\begin{figure}[b!]
    \centering
    \begin{subfigure}[b]{0.48\textwidth}
        \includegraphics[width=\textwidth]{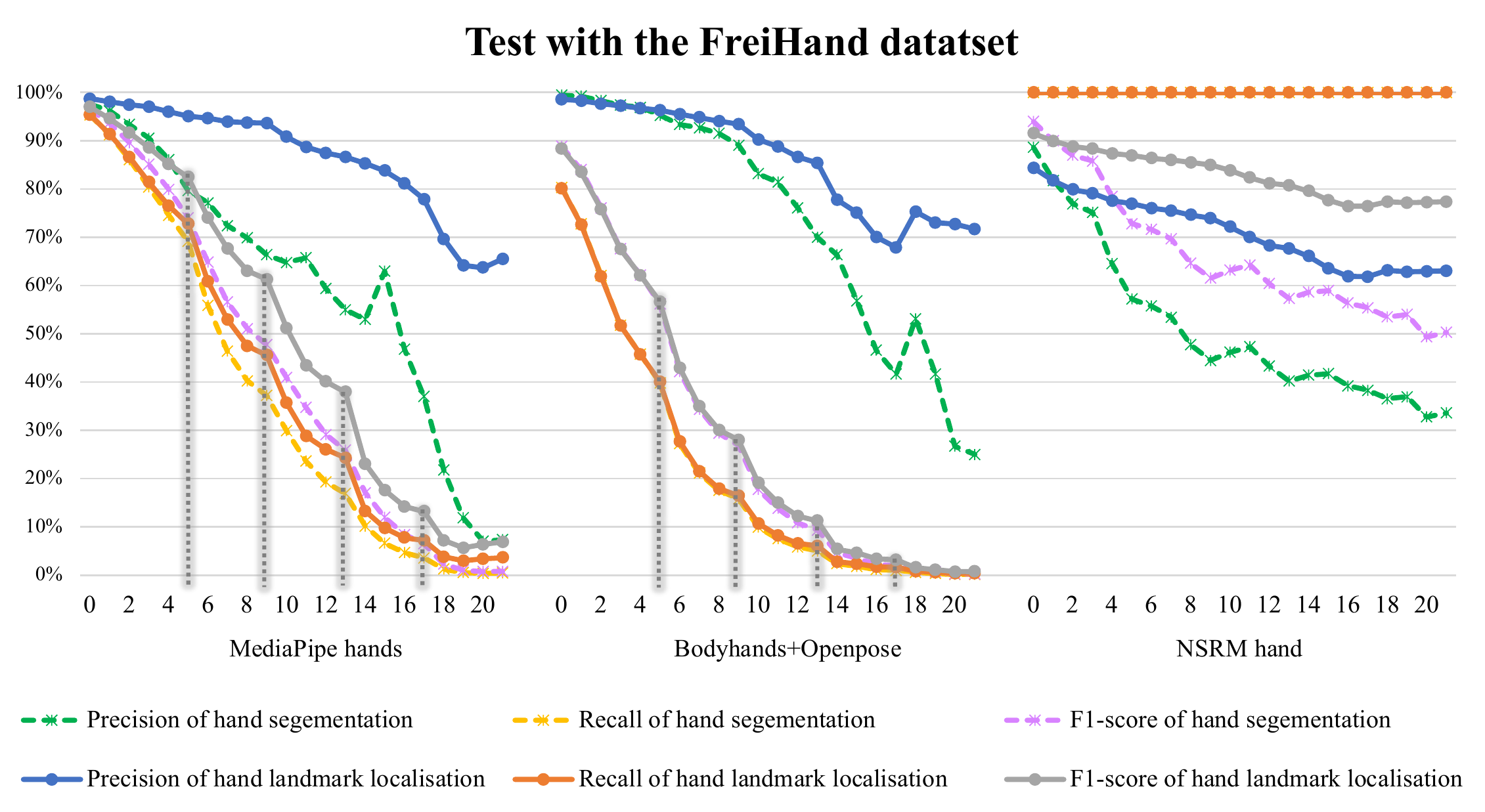}
    \end{subfigure}
    ~
    \begin{subfigure}[b]{0.48\textwidth}
        \includegraphics[width=\textwidth]{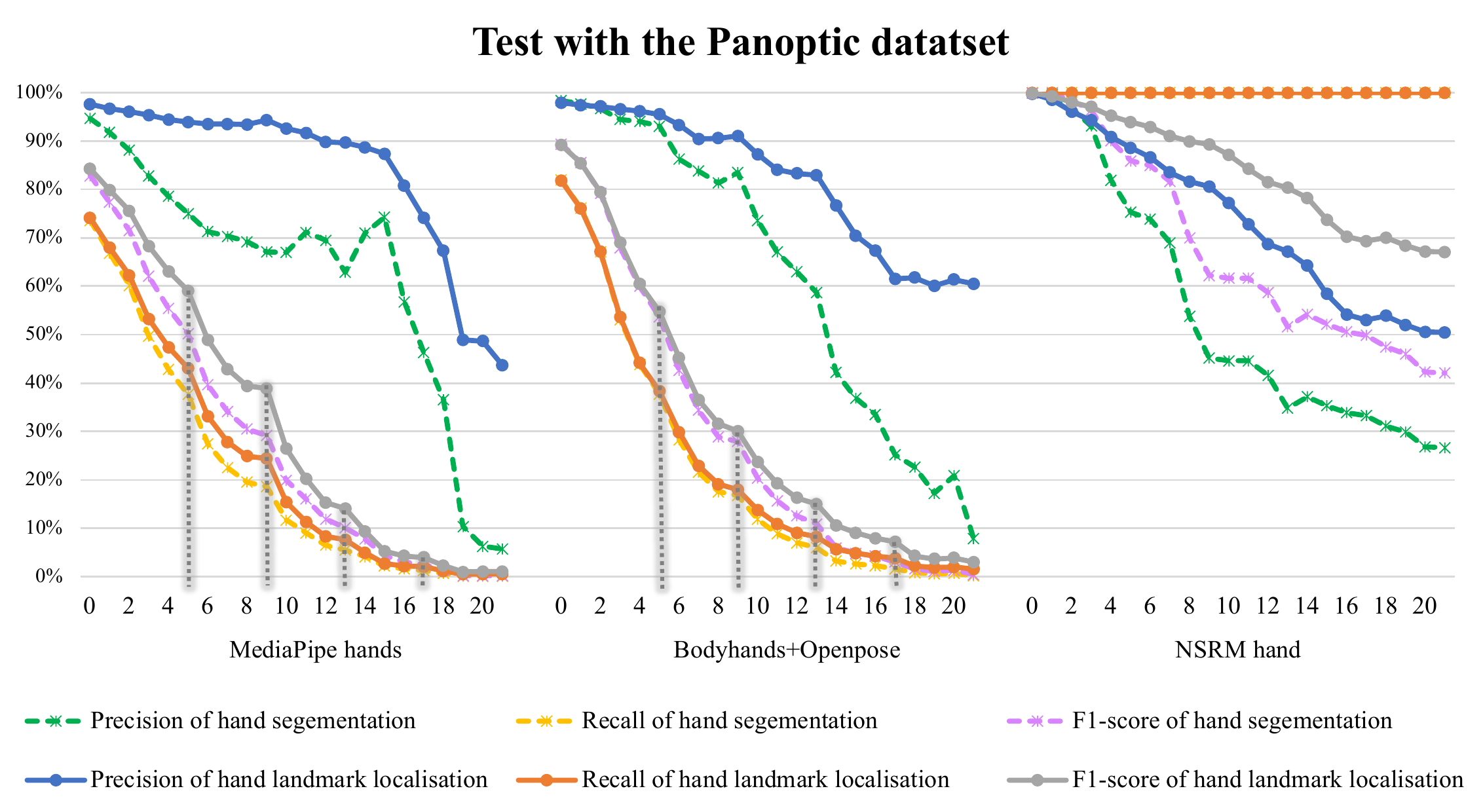}
    \end{subfigure}
    \caption{Results of MR{$_1$}. The vertical dot lines are used to label the crucial points. The experiment results obtained from the original images of hands without objects (baseline) are shown as the starting points of each sub-graph.}
    \label{fig:MR1_results}
\end{figure}

\textbf{However, the results shown in Figure \ref{fig:MR1_results} deviated from our expectations and reveal an inconsistency with MR{$_1$}. We can observe that the performance degradation of HPE models is not strongly associated with the increase of occluded key points. The occlusion of different key points results in varying levels of performance degradation, indicating that the 21 key points of the hand landmark are not equally important for hand pose estimation}.

In most of the cases involving TC{$_1$}, we found that the recall and F1-score decrease sharply at four crucial points, which are presented at level 5, level 9, level 13 and level 17 occlusions respectively and labelled by the vertical dot lines (see Figure \ref{fig:MR1_results}), where we finish occluding a specific finger. For example, level 5 occlusions cover the thumb from the original images, and level 13 occlusions cover the thumb, index, and middle fingers. Each time a finger is completely occluded, the shape of the hand is transformed, resulting in a rapid decline in the performance of these models. 

NSRM hand performs differently from the other two models, where the precision and F1-score fluctuate less, and each sub-graph of the model shows a continuously decreasing trend with respect to TC{$_1$}. In contrast to MediaPipe hands and BodyHands+Openpose, a slightly stronger correlation between the overall estimation performance and occlusion levels can be reflected by NSRM hand, leading the model to be less unfaithful to MR{$_1$}. 

% Besides, two more special points are identified at this stage, which present at the level 15 occlusion of MediaPipe hands and level 18 occlusion of BodyHands+Openpose, where we start to apply occlusions at the second lowest of joint of the ring finger occlusion (level 15 occlusion) and the lowest joints of the pinky finger (level 18 occlusion). Two local maximum precison is shown at these two points separately for MediaPipe hands and BodyHands+Openpose, and a sudden decrease in their precision occurs after we apply occlusions on the remaining key points. This finding reveal that 

% A local maximum precision value is shown at the level 15 occlusion in the first sub-graph for MediaPipe hands of Figure \ref{fig:MR1_results}, where we start to apply occlusions at the second lowest of joint of the ring finger, at which most people wear their wedding rings below the joint. Similarly, in the sub-graph for BodyHands+Openpose on the FreiHand dataset shown in Figure \ref{fig:MR1_results}, we can find a local maximum precision at the level 18 occlusion, where the lowest joints of the pinky finger are occluded. The findings reveal that these special key points, the 15th key point for MediaPipe Hands and the 18th key point for BodyHands+Openpose, may be important for a precise estimation,  

On the other hand, though the recall values of MediaPipe hands and BodyHands+Openpose suffers greatly from occlusions with greater fluctuation than NSRM hand, they are able to achieve a precision of approximately 90\% in localising hand landmarks for hands with up to level 10 occlusion. Surprisingly, the precision values of hand landmark localisation in neither of these two models are prone to zero, instead remaining above 40\% at the largest occlusion level, where the entire area of the hand is occluded. This may explain the underlying algorithms of MediaPipe hands and BodyHands+Openpose are designed to logically estimate hand poses within the segmented areas following the anatomy of hands. 

Moreover, the recall values of MediaPipe hands in both cases drop by around half from its baseline when tested with level 8 occlusion, while the recall values of BodyHands+Openpose drop by more than half from its baseline with level 5 occlusion on both datasets. This implies that the model begins to randomly estimate if a testing image contains a hand when the thumb and index fingers are partially occluded.

% The hand landmark localisation results of TC{$_1$} are illustrated in Figure \ref{fig:MR1_results}, where the patterns of their recall and F1-score appear to be similar to the patterns of their related hand segmentation results. By comparing with their respective TC{$_1$} hand segmentation results, the 15th key point for MediaPipe Hands and the 18th key point for BodyHands+Openpose are less important here, and the precision of these three models for hand landmark localisation declines in a comparatively slow manner. 

The above observation found via MR{$_1$} could imply a possibility as the cause for such irregularity: The significance of each finger is different from each other. We then examine our second metamorphic relation MR{$_2$} through test cases TC{$_2$}-TC{$_6$}, which were designed to see how these models respond to hands with occlusion applied on the individual finger. Figure \ref{fig:MR2_samples} illustrates samples of images generated to verify MR{$_2$}. 

\begin{figure}[t]
    \centering
    \begin{subfigure}[b]{0.14\textwidth}
        \caption*{Index finger occlusion (TC{$_3$})}
        \includegraphics[width=\textwidth]{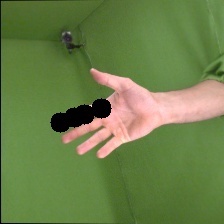}
    \end{subfigure}
    ~
    \begin{subfigure}[b]{0.14\textwidth}
        \caption*{Middle finger occlusion (TC{$_4$})}
        \includegraphics[width=\textwidth]{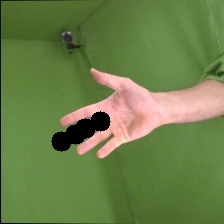}
    \end{subfigure}
    ~
    \begin{subfigure}[b]{0.14\textwidth}
        \caption*{Ring finger occlusion (TC{$_5$})}
        \includegraphics[width=\textwidth]{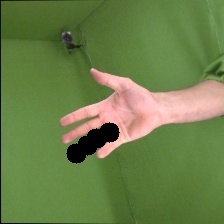}
    \end{subfigure}
    \caption{Samples of images for MR{$_2$}}
    \label{fig:MR2_samples}
\end{figure}

\begin{figure}[b!]
    \centering
    \begin{subfigure}[b]{0.48\textwidth}
        \includegraphics[width=\textwidth]{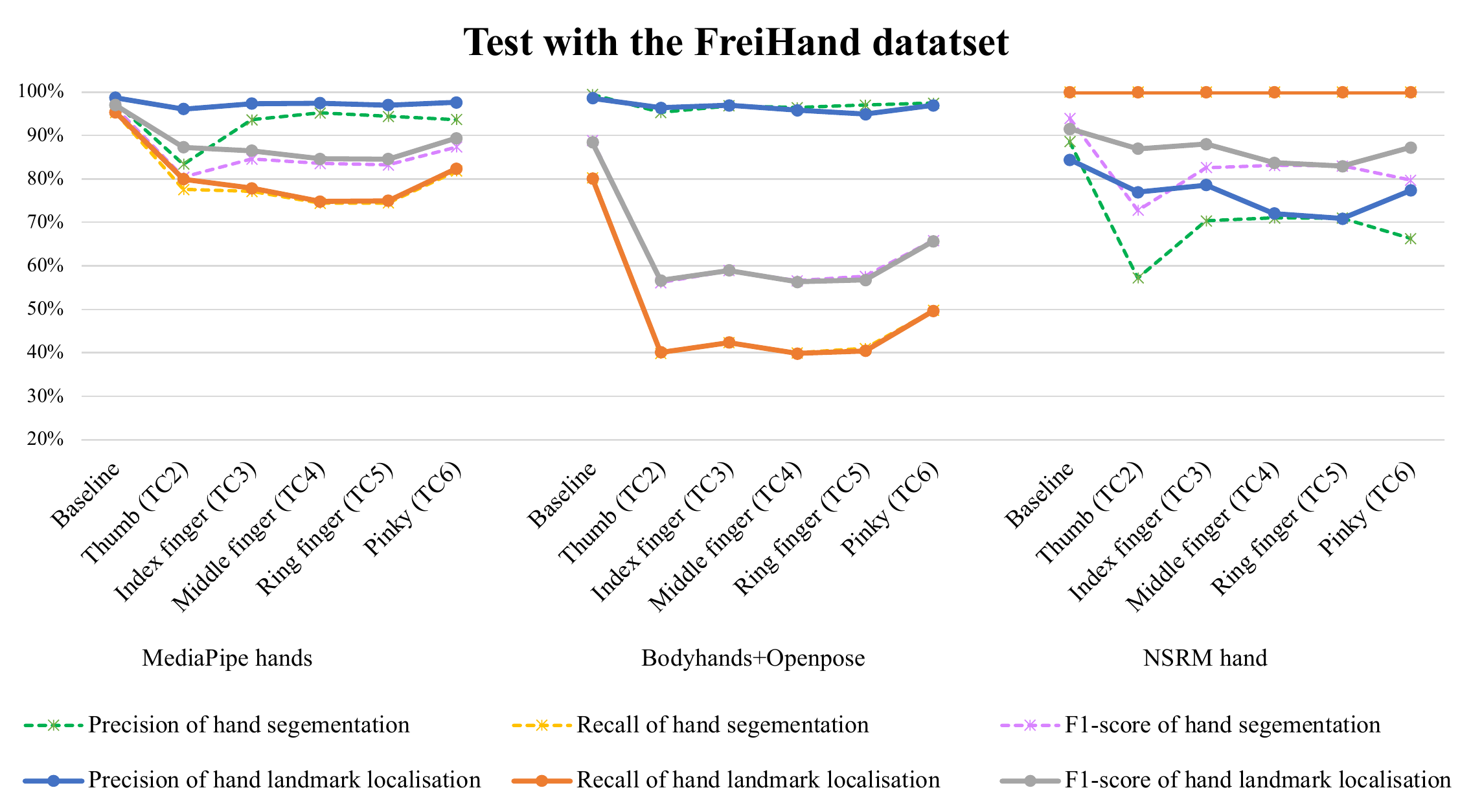}
    \end{subfigure}
    ~
    \begin{subfigure}[b]{0.48\textwidth}
        \includegraphics[width=\textwidth]{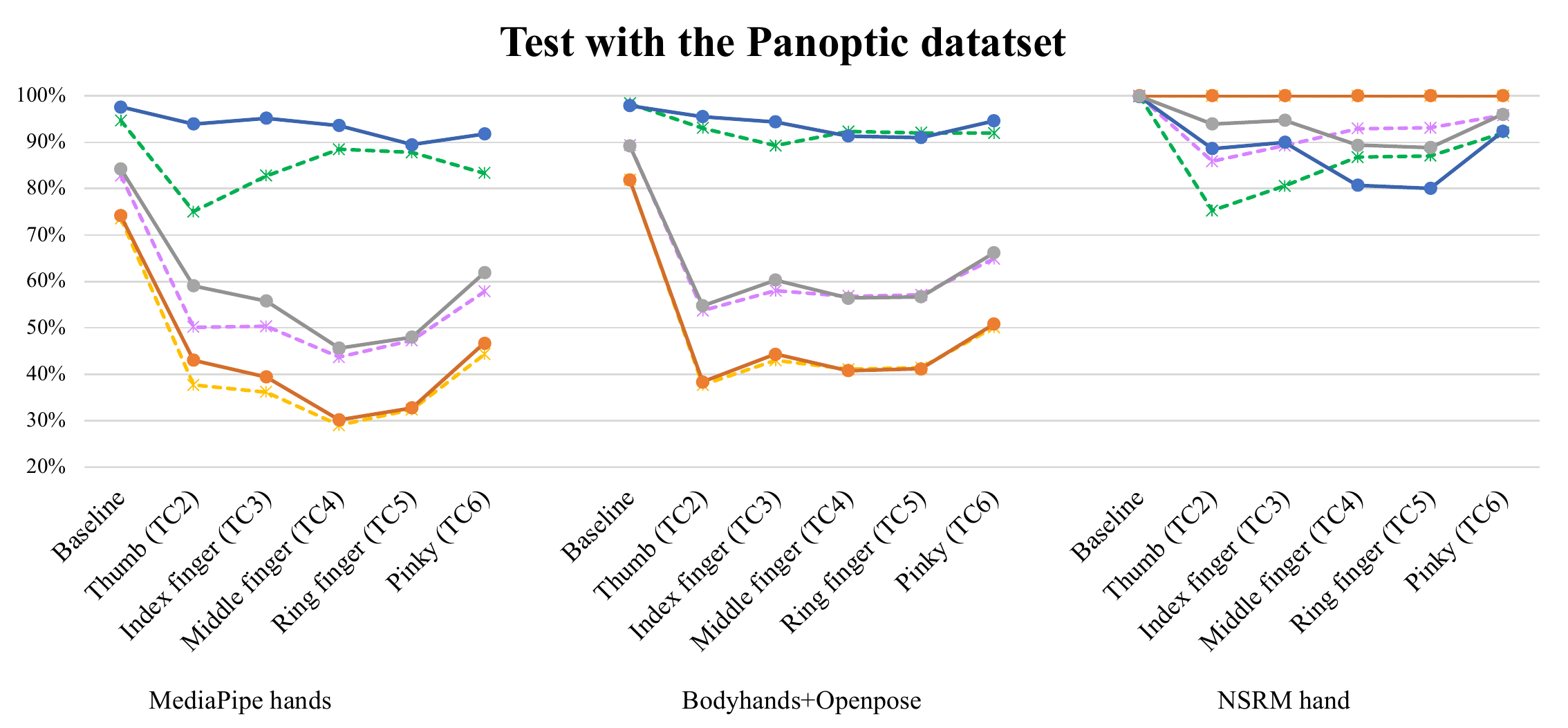}
    \end{subfigure}
    \caption{Results of MR{$_2$}}
    \label{fig:MR2_results}
\end{figure}

We expect that the performance of these models will not be degraded by the introduction of slight occlusion on hand regions that are prone to be occluded, such as a particular finger. \textbf{While the results shown in Figure \ref{fig:MR2_results} were inconsistent with MR{$_2$}, we observed that applying occlusions on the thumb fingers generally causes these models to perform less precisely than applying on other fingers.} Additionally, the recall values of MediaPipe hands and BodyHands+Openpose are always lower than the matched precision values, which implies that these two models follow a relatively restrictive criterion for identifying objects as hands. Hence, applying occlusions on the thumb will hugely transform the shape of hands, resulting in their recall values drop by roughly 20\% (MediaPipe hands on the FreiHand dataset), and around 40\% (MediaPipe hands on the Panoptic dataset, BodyHands+Openpose on both datasets) from their responding baseline. The worst precision can be observed from the results of the NSRM hand, which is unlike the precision values of MediaPipe hands and BodyHands+Openpose that consistently hover above 75\%. We concluded that applying occlusions to either of the fingers causes greater degradation in the precision of NSRM hand, especially on the FreiHand dataset. 

Moving on from occlusions, the impact of illumination variations can also be attributed to hand regions. The third metamorphic relation, MR{$_3$}, is therefore utilised to explore this impact. Its corresponding test cases TC{$_7$}-TC{$_{10}$} (refer to Figure \ref{fig:MR3_samples}) are derived to observe how would these test cases impact the selected models with respect to strong underexposure TC{$_7$}, underexposure TC{$_8$}, overexposure TC{$_9$}, and strong overexposure TC{$_{10}$}. The gamma values used to change the exposure rate of the original image are selected at 5, 2, 0.5, and 0.2 for generating TC{$_7$},  TC{$_8$},  TC{$_9$},  and TC{$_{10}$} respectively. We can see from Figure \ref{fig:MR3_samples} that the hands contained in these images are of clear shapes after these gamma values are applied to change the exposure rates. 

\begin{figure}[b!]
    \centering
    \captionsetup{justification=centering}
    \begin{subfigure}[b]{0.14\textwidth}
        \caption*{Strong underexposure (TC{$_7$})}
        \includegraphics[width=\textwidth]{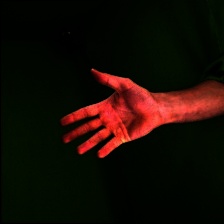}
    \end{subfigure}
    \hspace{3em}% Space between image B and C
    \begin{subfigure}[b]{0.14\textwidth}
        \caption*{Underexposure (TC{$_8$})}
        \includegraphics[width=\textwidth]{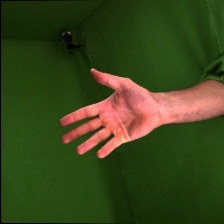}
    \end{subfigure}
    \\
    \begin{subfigure}[b]{0.14\textwidth}
        \caption*{Overexposure (TC{$_9$})}
        \includegraphics[width=\textwidth]{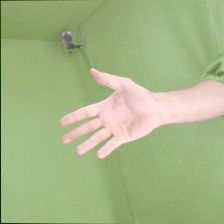}
    \end{subfigure}
    \hspace{3em}% Space between image B and C
    \begin{subfigure}[b]{0.14\textwidth}
        \caption*{Strong overexposure (TC{$_{10}$})}
        \includegraphics[width=\textwidth]{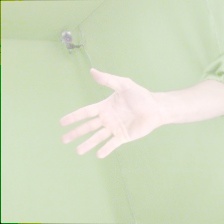}
    \end{subfigure}
    \caption{Samples of images for MR{$_3$}}
    \label{fig:MR3_samples}
\end{figure}

\begin{figure}[h!]
    \centering
    \begin{subfigure}[b]{0.48\textwidth}
        \includegraphics[width=\textwidth]{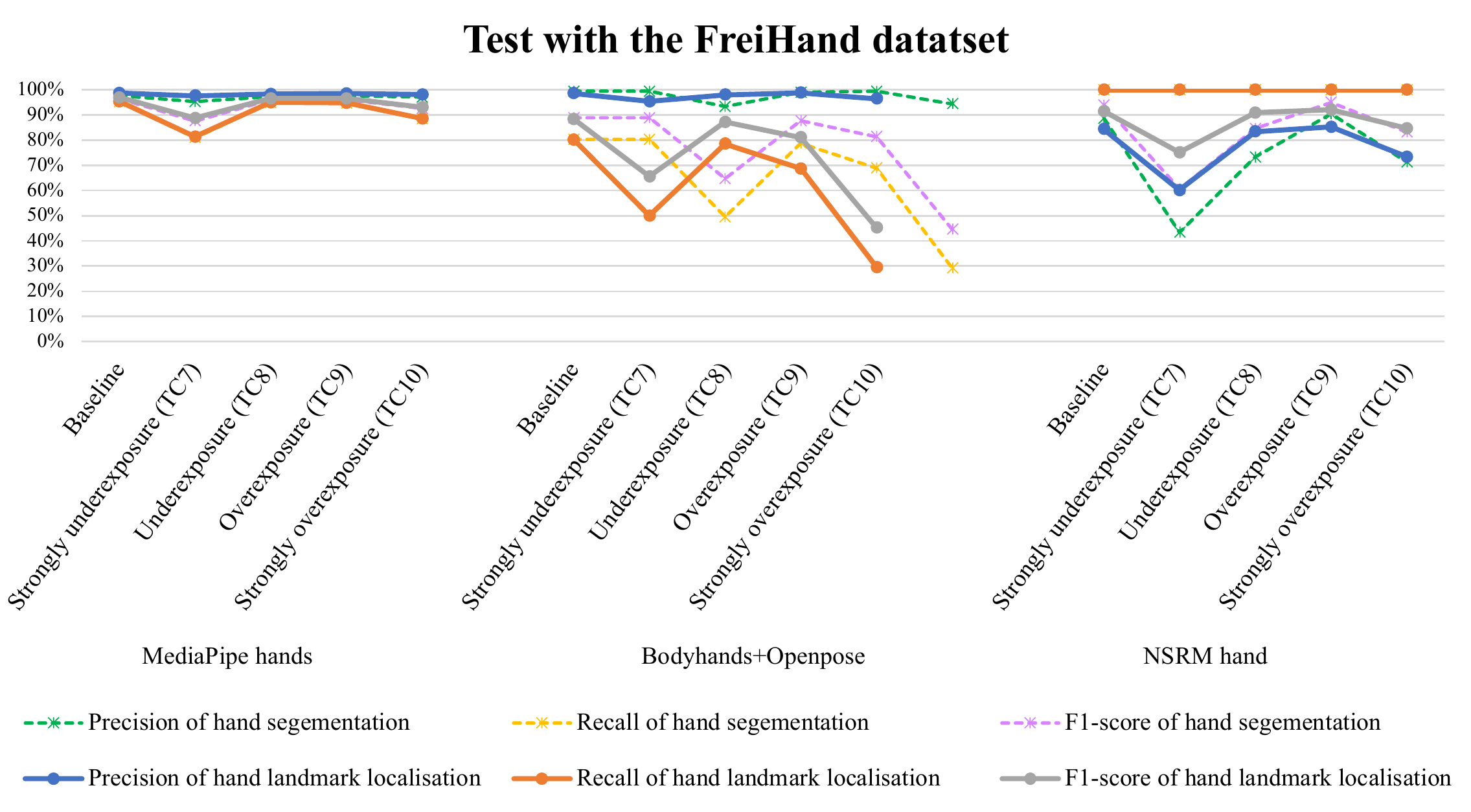}
    \end{subfigure}
    ~
    \begin{subfigure}[b]{0.48\textwidth}
        \includegraphics[width=\textwidth]{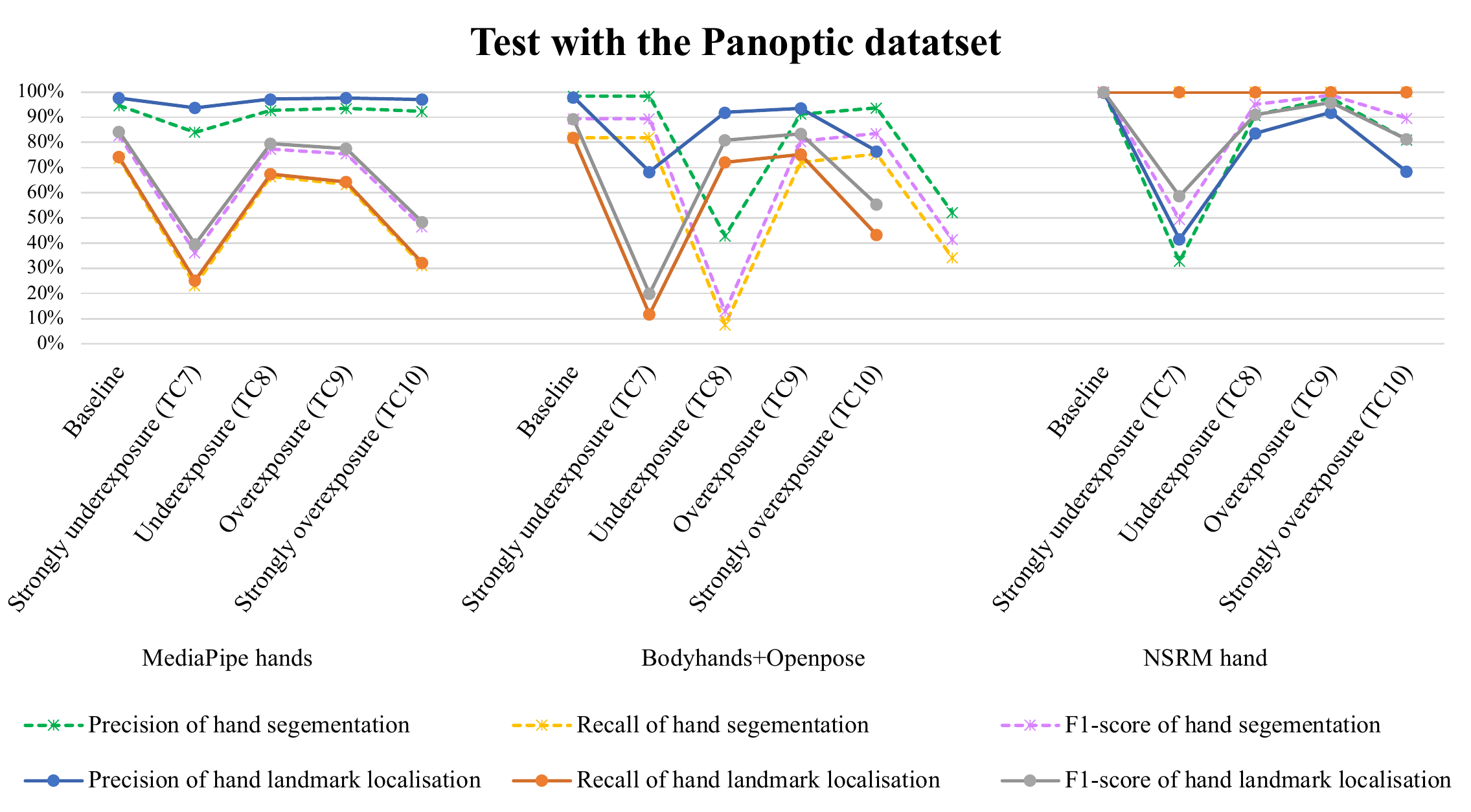}
    \end{subfigure}
    \caption{Results of MR{$_3$}}
    \label{fig:MR3_results}
\end{figure}

As stated in MR{$_3$}, we expect that if the exposure rate of the original image is changed to mimic the illumination of different operating environments, the performance of HPE models should not be degraded by a huge margin. \textbf{However, all of these models show different degrees of performance degradation under these four test cases (see Figure \ref{fig:MR3_results}), which is found to be unfaithful to MR{$_3$}}. TC{$_7$} and TC{$_{10}$} post the most severe influence on the models' performance than TC{$_8$} and TC{$_9$}. On average, NSRM hand fails to correctly estimate more than half of underexposed hands from TC{$_7$}, and the recall values of BodyHands+Openpose decrease by more than 39 \% on TC{$_{10}$}. One interesting finding is that the performance of MediaPipe hands is relatively robust when tested with the test cases generated from the FreiHand dataset, which could imply that the model may have experience in handling the relevant illumination variations during its training processing, but its experience may not be enough to deal with all types of illumination variations. We now have reason to believe that these models are not familiar with illumination variations, and they will not function normally for hands that are strongly underexposed or strongly overexposed. 

\begin{figure}[b!]
    \centering
    \captionsetup{justification=centering}
    \begin{subfigure}[b]{0.14\textwidth}
        \caption*{Vertical motion \\ blur (TC{$_{11}$})}
        \includegraphics[width=\textwidth]{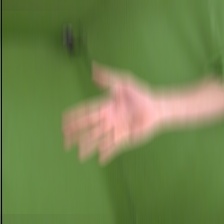}
    \end{subfigure}
    ~
    \begin{subfigure}[b]{0.14\textwidth}
        \caption*{Horizontal motion blur (TC{$_{12}$})}
        \includegraphics[width=\textwidth]{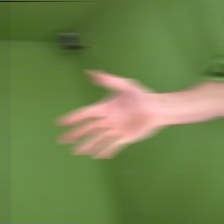}
    \end{subfigure}
    ~
    \begin{subfigure}[b]{0.14\textwidth}
        \caption*{Diagonal motion blur (TC{$_{13}$})}
        \includegraphics[width=\textwidth]{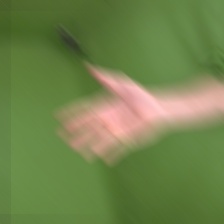}
    \end{subfigure}
    \caption{Samples of images for MR{$_4$}}
    \label{fig:MR4_samples}
\end{figure}

Next, we are interested to verify MR{$_4$} that the performance of HPE models should not be degraded by the introduction of motion blur. We expect some tolerance of the selected HPE model with regard to motion blur, as it is very common for motion blur to be involved in fast hand movements to gesture and play sign language. To validate MR{$_4$}, we then generate three types of motion blur (see Figure \ref{fig:MR4_samples}), involving vertical motion blur (TC{$_{11}$}), horizontal motion blur (TC{$_{12}$}), and diagonal motion blur (TC{$_{13}$}). 

However, \textbf{the results in Figure \ref{fig:MR4_results} were found to be inconsistent with MR{$_4$}, and none of these models exhibits a robust performance against motion blur.} As observed in Figure \ref{fig:MR4_results}, the precision of both MediaPipe hands and BodyHands+Openpose drop by a similar amount compared to the baseline, though they are all above 87\%. As opposed to the precision, their recall values differ a lot. BodyHands+Openpose's recall values on each test case of MR{$_4$} is prone to below 50\%, and the model is severely affected by vertical and diagonal motion blur, resulting in a dramatic decrease in the number of identified hand poses and their corresponding key points. A slightly better result can be found from MediaPipe hands, except for the results obtained from TC{$_{13}$}, where the diagonal motion blur still poses the greatest effect on its performance, MediaPipe hands' recall values are all above 50 \%. This transformation in the recall values indicates that motion blur managed to confuse the two models to consider a huge number of hands as other objects, which further unveils the domain distribution discrepancies issue and the unstable performance of these models. On the other hand, we can see NSRM hand shows a greater fluctuation in precision than recall due to its design to assume the involvement of a hand in any image. However, this property cause NSRM hands to be unable to filter out those data that the model is not capable to deal with, which then reveals its design defect when facing unfamiliar domains. 

\begin{figure}[h!]
    \centering
    \begin{subfigure}[b]{0.48\textwidth}
        \includegraphics[width=\textwidth]{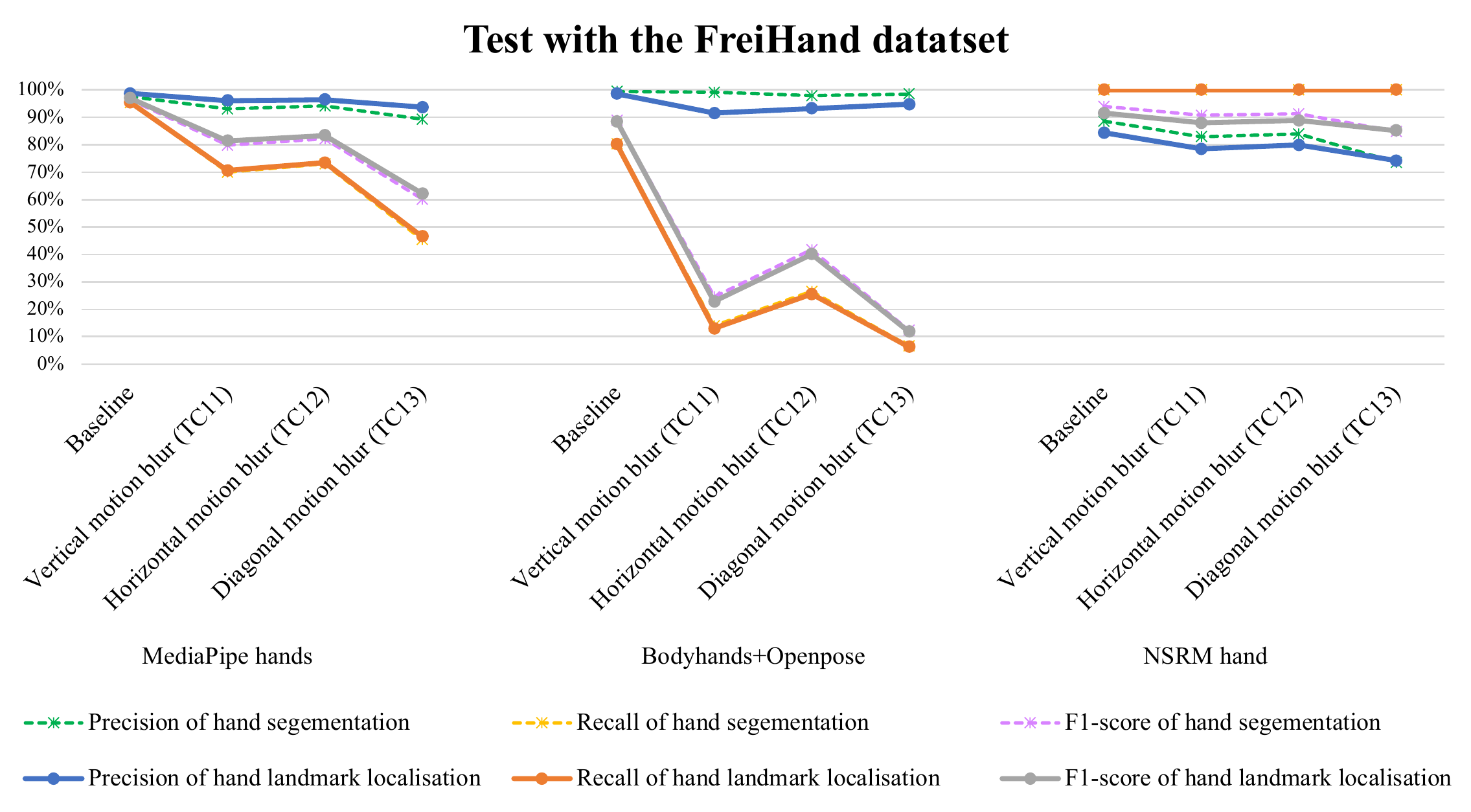}
    \end{subfigure}
    ~
    \begin{subfigure}[b]{0.48\textwidth}
        \includegraphics[width=\textwidth]{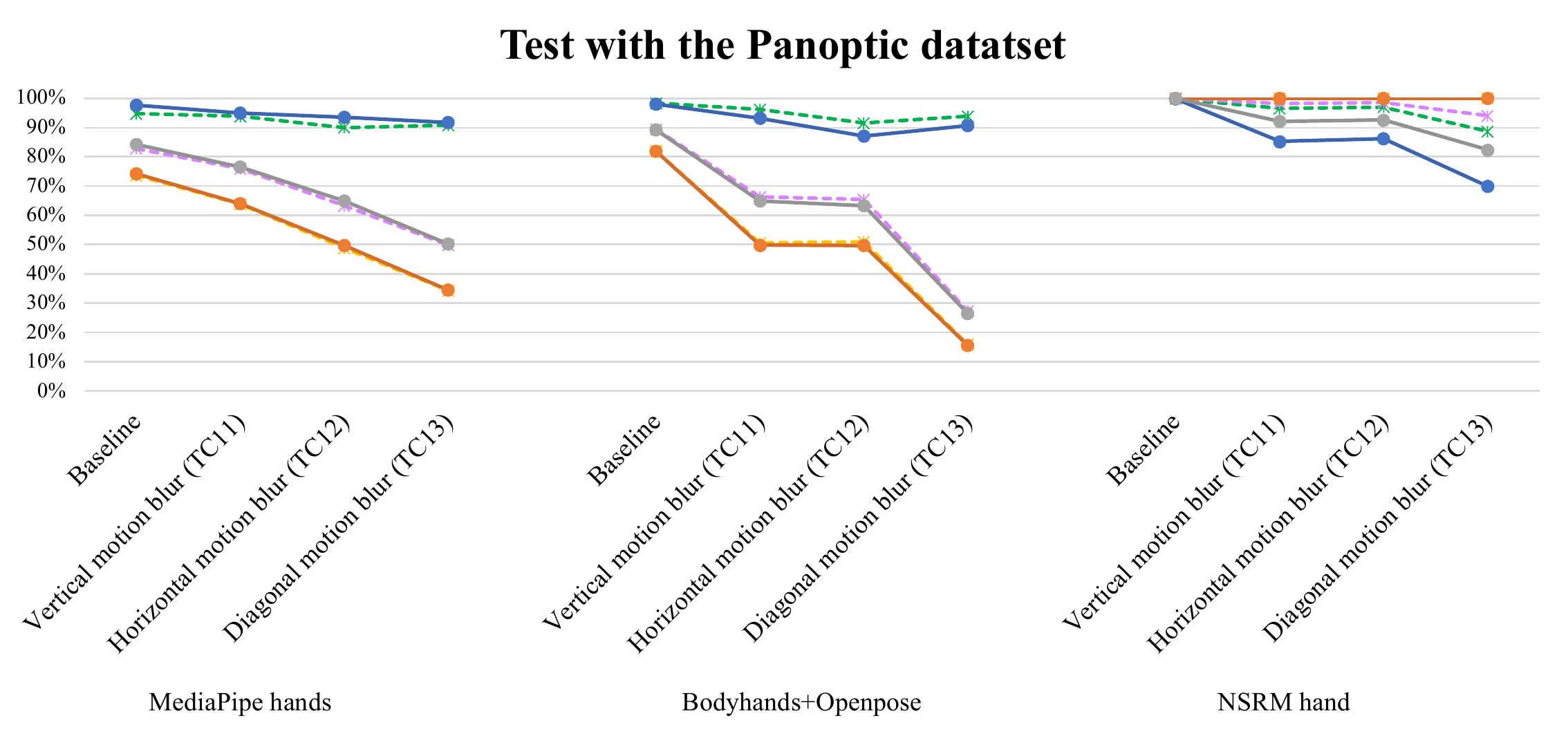}
    \end{subfigure}
    \caption{Results of MR{$_4$}}
    \label{fig:MR4_results}
\end{figure}

Finally, we review the results of our robustness analysis for all test cases. Generally, we found that MediaPipe hands and BodyHands+Openpose perform at their worst with test cases involving diagonal motion blur, while NSRM hand is more susceptible to illumination variations, especially dark environments. Based on these results, 
MediaPipe hands and BodyHands+Openpose exhibit high precision, making them relatively suitable for applications involving gesture cognition or sign language recognition, despite the demands of improvements in their robustness. On the other hand, NSRM hand demonstrates high sensitivity, which may ensure its application in virtual or augmented reality under in-door environments where light conditions can be controlled, hands are less likely to be occluded by objects, and users are more likely to learn control strategies. With more variations from different aspects wisely considered in the training stage of these models, we believe they are able to seek ways to overcome the current operating constraints caused by occlusions, illumination, and motion blur. 

\section{Conclusion}
In this paper, we present a method to evaluate the robustness of state-of-the-art hand pose estimation (HPE) models via metamorphic testing (MT). Leveraging on the original datasets (which contain pristine hand images), MT enables us to introduce variations between the hand samples and create new datasets\footnote{\url{https://github.com/mpuu00001/Robustness-Evaluation-in-Hand-Pose-Estimation}} to mimic real operating environments, which largely reduces the time and effort to manually curate such large scale datasets. We believe that the dataset that we curated (which contain variations from the four metamorphic relationships) enable researchers to construct and evaluate their proposed HPE models with a wider domain distribution. 

By investigating the impact of occlusions, illumination variations, and motion blur, we discovered that they are the main considerations to the poor performance of the selected HPE models. All the models show different advantages and disadvantages in estimating hand poses under settings involving these three operating constraints, where MediaPipe hands and BodyHands+Openpose tend to identify less number of hands with acceptable precision. On the other hand, NSRM hand will always produce estimation results with comparatively fluctuated precision. Our experimental findings prove that 1) MT is fairly effective in identifying defects that could cause models to fail in unintuitive ways; 2) existing HPE models are not robust against occlusions, illumination variations, and motion blur. Our findings show that the impact of these three constraints not only degrades the precision of HPE models, but also decreases the number of hands to be identified. We believe that this study can assist to spur new research opportunities and ideas in the field of hand pose estimation and its applications.

\printbibliography 
\end{document}